\documentclass[runningheads]{llncs}
\usepackage{todonotes}
\usepackage{graphicx}
\usepackage{booktabs}
\usepackage{hyperref}
\usepackage{subcaption}

\usepackage[activate={true,nocompatibility},final,tracking=true,kerning=true,spacing=true,factor=1100,stretch=10,shrink=10]{microtype}

\begin{document}
\title{Dutch Humor Detection\\ by Generating Negative Examples}
\author{Thomas Winters\orcidID{0000-0001-7494-2453} \and
Pieter Delobelle\orcidID{0000-0001-5911-5310}}
\authorrunning{T. Winters, P. Delobelle}
\institute{
Dept. of Computer Science; Leuven.AI \\
KU Leuven, Belgium \\
\email{firstname.lastname@kuleuven.be}
}
\maketitle              %
\begin{abstract}

Detecting if a text is humorous is a hard task to do computationally, as it usually requires linguistic and common sense insights.
In machine learning, humor detection is usually modeled as a binary classification task, trained to predict if the given text is a joke or another type of text.
Rather than using completely different non-humorous texts, we propose using text generation algorithms for imitating the original joke dataset to increase the difficulty for the learning algorithm.
We constructed several different joke and non-joke datasets to test the humor detection abilities of different language technologies.
In particular, we compare the humor detection capabilities of classic neural network approaches with the state-of-the-art Dutch language model RobBERT.
In doing so, we create and compare the first Dutch humor detection systems.
We found that while other models perform well when the non-jokes came from completely different domains, RobBERT was the only one that was able to distinguish jokes from generated negative examples.
This performance illustrates the usefulness of using text generation to create negative datasets for humor recognition, and also shows that transformer models are a large step forward in humor detection.

\keywords{Computational Humor \and Humor Detection \and RobBERT \and BERT model }
\end{abstract}
\section{Introduction}

Humor is an intrinsically human trait.
All human cultures have created some form of humorous artifacts for making others laugh \cite{caron2002humorevolution}. 
Most humor theories also define humor in function of the reaction of the perceiving humans to humorous artifacts.
For example, according to the popular incongruity-resolution theory, we laugh because our mind discovers that an initial mental image of a particular text is incorrect, and that this text has a second, latent interpretation that only becomes apparent when the punchline is heard \cite{suls1972two,ritchie1999irtheory}.
To determine that something is a joke, the listener thus has to mentally represent the set-up, followed by detecting an incongruity caused by hearing the punchline, and resolve this by inhibiting the first, non-humorous interpretation and understanding the second interpretation \cite{deckers1990humor,gibson2016good}.
Such humor definitions thus tie humor to the abilities and limitations of the human mind: if the joke is too easy or too hard for our brain, one of the mental images might not get established, and lead to the joke not being perceived as humorous.
As such, making computers truly and completely recognize and understand a joke would not only require the computer to understand and notice the two possible interpretations, but also that a human would perceive these as two distinct interpretations.
Since the field of artificial intelligence is currently nowhere near such mental image processing capacity, truly computationally understanding arbitrary jokes seems far off.

While truly understanding jokes in a computational way is a challenging natural language processing task, there have been several studies that researched and developed humor detection systems \cite{taylor2004knockknock,mihalcea2005humorrecognition,kiddon2011deviant,cattle2018anchor,van2018homonym,weller2019transformer,annamoradnejad2020colbert}.
Such systems usually model humor detection as a binary classification task where the system predicts if the given text is a joke or not.
The non-jokes often come from completely different datasets, such as news %
and proverbs %
\cite{mihalcea2005humorrecognition,yang2015humor,cattle2018anchor,van2018homonym,chen2018humor,annamoradnejad2020colbert}.
In this paper, we create the non-joke dataset by using text generation algorithms designed to mimic the original joke dataset by only using words that are used in the joke corpus \cite{winters2019torfsbot}.
This dataset thus substantially increases the difficulty of humor detection, especially for algorithms that use word-based features, given that coherence plays a more important role in distinguishing the two.
We use the recent RobBERT model to test if its linguistic abilities allow it to also tackle the difficult challenge of humor detection, especially on our new type of dataset.
As far as the authors are aware, this paper also introduces the first Dutch humor detection systems.

\section{Background}

\subsection{Neural Language Models}

Neural networks perform incredibly well when dealing with a fixed number of features.
When dealing with sequences of varying lengths, recurrent connections to previous states can be added to the network, as done in recurrent neural networks (RNN).
Long short-term memory (LSTM) networks are a variety of RNN that add several gates for accessing and forgetting previously seen information.
This way, a sequence can be represented by a fixed-length feature vector by using the last hidden states of multiple LSTM cells~\cite{schmidhuber1997lstm}.
Alternatively, if a maximum sequence length is known, the input size of a neural network could be set to this maximum, and e.g. allow for using a convolutional neural network (CNN).

Entering text into a recurrent neural network is usually done by processing the text as a sequence of words or tokens,
each represented by a single vector from pre-trained embeddings containing semantic information~\cite{mikolov2013word2vec}.
These vectors are obtained from large corpora in the target language, where the context of a token is predicted e.g. using Bag-of-Words (BOW)~\cite{mikolov2013word2vec}. %

\subsubsection{BERT}

The BERT model is a powerful language model that improved many state-of-the-art performances on NLP tasks \cite{devlinBERT2019a}.
It is built using a transformer encoder stack consisting of self-attention heads to create a bidirectional language model \cite{vaswani2017attention}.
These attention mechanisms allow BERT to distinguish different meanings for particular words based on the context by using contextualized embeddings.
For example, even though the word \emph{``stick''} could be both a noun as well as a verb, normal word embeddings assign the same vector to both meanings.

BERT is trained in a self-supervised way by predicting missing words in sentences, and predicting if two randomly chosen sentences are subsequent or not.
After this pre-training phase, additional heads can be fine-tuned on particular datasets to classify full sentences, or to classify every token of a sentence.
The model exhibits large quantities of linguistic knowledge (e.g. for resolving coreferences, POS tagging, sentiment analysis) and achieved state-of-the-art performance on many different language tasks.
This model later got critically re-evaluated and improved in the RoBERTa model, which uses a revised training regime \cite{liu2019roberta}. %

\subsubsection{RobBERT}

RobBERT \cite{delobelle2020robbert} is a recent Dutch language model trained using the RoBERTa regime \cite{liu2019roberta} on the Dutch OSCAR corpus section \cite{ortizsuarez2019oscar}.
Like RoBERTa, RobBERT also outperforms other Dutch language models in a wide range of complex NLP tasks e.g. coreference resolution and sentiment analysis \cite{delobelle2020robbert}.

\subsection{Humor Detection}

Humor is not an easy feat for computational models.
True humor understanding would need large quantities of linguistic knowledge and common sense about the world to know that an initial interpretation is being revealed to be incompatible with the second, hidden interpretation fitting the whole joke rather than only the premise.
Many humor detection systems use hand-crafted (often word-based) features to distinguish jokes from non-jokes \cite{taylor2004knockknock,mihalcea2005humorrecognition,kiddon2011deviant,van2018homonym}. %
Such word-based features perform well when the non-joke dataset is using completely different words than the joke dataset.
From humor theory, we know that the order of words matter, since stating the punchline before the setup would only cause the second interpretation of the joke to be discovered, making the joke lose its humorous aspect \cite{ritchie1999irtheory}.
Since word-based humor detectors often fail to capture such temporal differences, more contextual-aware language models are required to capture the true differences between jokes and non-jokes.

Using a large pre-trained model like the recent BERT-like models is thus an interesting fit for the humor detection task.
One possible downside is that these models are not well suited for grasping complex wordplay, as their tokens are unaware of relative morphological similarities, due to the models being unaware of the letters of the tokens \cite{branwen2020gpt}.
Nevertheless, BERT-like models have performed well on English humor recognition datasets \cite{weller2019transformer,annamoradnejad2020colbert}.

Recently, several parallel English satirical headline corpora have been released for detecting humor, which might help capture subtle textual differences that create or remove humor \cite{west2019reverse,hossain2019president}.
Lower resource languages however usually do not have access to such annotated parallel corpora for niche tasks like humor detection.
While there has been some Dutch computational humor research \cite{winters2019torfsbot,winters2019clin,winters2019samsonbot}, there has not been any published research about Dutch humor detection, nor are there any public Dutch humor detection data sets available.

\subsection{Text Generation for Imitation}

There are many types of text generation algorithms.
The most popular type of text generation algorithms use statistical models to iteratively predict the next word given previous words, e.g. n-gram based Markov models or GPT-2 and GPT-3 \cite{radford2019gpt2,brown2020gpt3}. These algorithms usually generate locally coherent text \cite{winters2019torfsbot}.
A common downside is that the output tends to have globally different structures than the training data (e.g. much longer or shorter), or even break (possibly latent) templates of the dataset \cite{winters2020gitta}.

\subsubsection{Dynamic Templates}

Templates can be seen as texts with holes, which are later filled, and thus enforce a global textual structure.
One approach for learning these is the dynamic template algorithm (DT), which is designed to replace context words with other, grammatically similar words \cite{winters2019torfsbot}.
It achieves this by analyzing the part-of-speech (POS) tags in the dynamic template text and replaces these words with context words with the same POS tags.
It prioritizes low unigram-frequency words, as these are usually key words determining the context of the text.
This way, the dynamic template algorithm generates a large variety of more nonsensical versions of given texts, using only words from the corpus.

\section{Data}

\subsection{Collecting Datasets}

We collected a Dutch joke dataset by combining the jokes found on Kidsweek\footnote{\url{https://www.kidsweek.nl/moppen}}, DeBesteMoppen\footnote{ \url{https://www.debestemoppen.nl/}} and LachJeKrom\footnote{\url{https://www.lachjekrom.com/}}.
This resulted in a dataset of \textbf{3235 jokes}.

For the non-joke datasets, we first collected several datasets inspired by the type of datasets used in English humor detection, namely proverbs and news
 \cite{mihalcea2005humorrecognition,yang2015humor,cattle2018anchor,van2018homonym,chen2018humor,weller2019transformer,annamoradnejad2020colbert}.
The proverbs dataset originates from the Dutch proverbs Wikipedia page and contains \textbf{1887 proverbs}.
The news dataset are \textbf{3235 headlines} uniformly sampled from the 100K Dutch news headlines dataset \cite{yeh2019dpgmedia2019}.

\subsection{Negative Generation: Generating Non-Jokes from Jokes}\label{ss:generating}

Since news and proverbs use completely different words and structures, there is a need for a new type of challenging dataset for humor recognition that uses non-jokes that are close to jokes.
Given the fragile nature of a joke, changing several important words usually turn the joke into a non-humorous text.
We propose a new type of dataset for humor detection by generating negative examples by automatically imitating the joke dataset.
The dynamic template algorithm is a right fit for this, as it will not change the global structure like Markov models might do and is less prone to plagiarising large parts of the training corpus \cite{winters2019torfsbot}.
The DT algorithm creates absurd, but globally similar texts, by grammatically similar words into another joke.
For example, the joke
\emph{``Wat is groen en plakt aan de muur? Kermit de sticker!''}\footnote{\emph{``What's green and adheres to the wall? Kermit the Sticker''}, pun on \emph{``kikker''} (\emph{``frog''}) }
was turned into the non-joke
\emph{``Wat is groen en telefoneert aan de muur? Kermit de spin!''}\footnote{\emph{``What's green and telephones on the wall? Kermit the Spider''}}.

We chose the same parametrisation used in the original paper (see Appendix~\ref{app:dt}) \cite{winters2019torfsbot}.
The resulting non-jokes thus only use words from the jokes dataset, with comparable frequencies, and still have similar grammatical structures, albeit nonsensical content.
This way, language classifiers that just learn which words are more common in jokes (e.g. \emph{``oen''},  \emph{``Jantje''},  \emph{``blond''}...) will be at a disadvantage compared to models that have better insight in the semantic coherence of a joke.
Another advantage of this method for parallel corpus creation is that it is easily extensible to other lower resource languages.

\section{Evaluation}

We devised two types of learning tasks for detecting humor in these new datasets.
The first is the classic humor detection task with binary labels representing joke and non-joke.
The second is a pairwise humor detection task, where given a joke and a non-joke, the algorithm needs to detect which of the two is a joke.

\subsection{Models}

We compare four different models\footnote{The code, models, data collectors and demo are available on \url{https://github.com/twinters/dutch-humor-detection}.}, namely
a Naive Bayes classifier with the TF-IDF of 3000 (1,3)-grams as features,
an LSTM with Dutch word embeddings~\cite{tulkens2016evaluating},
a CNN with two convolutional layers and max pooling on Dutch word embeddings~\cite{kim2014convolutional},
and
RobBERT~\cite{delobelle2020robbert}.
The use of LSTMs and CNNs allows us to compare the RobBERT model with the previous generation of neural language models.

\subsection{Classification Experiment}

\begin{figure}[tb]
\begin{subfigure}{.245\textwidth}
  \centering
  \includegraphics[width=\linewidth]{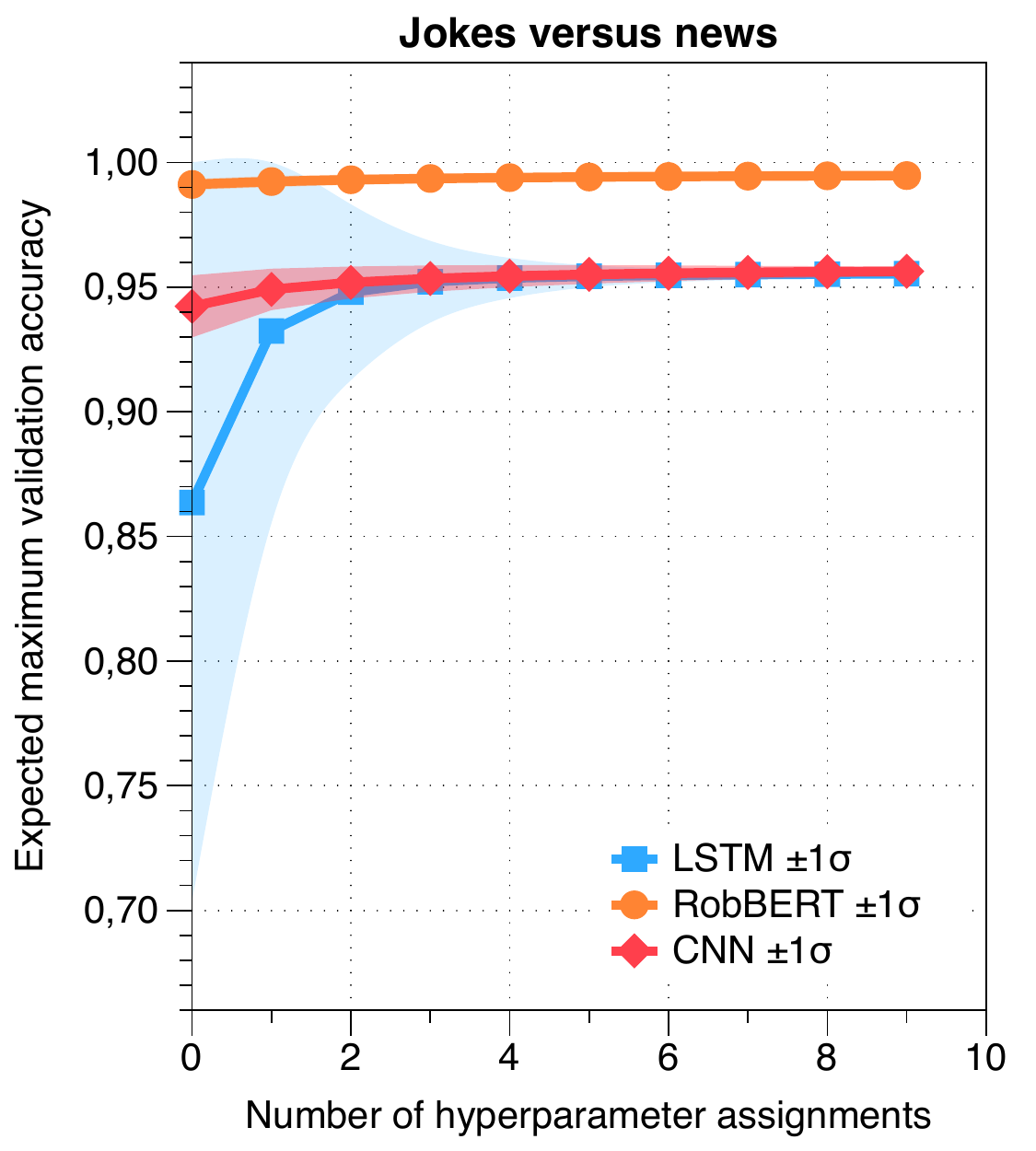}  
  \caption{News}
  \label{fig:sub-a}
\end{subfigure}
\begin{subfigure}{.245\textwidth}
  \centering
  \includegraphics[width=\linewidth]{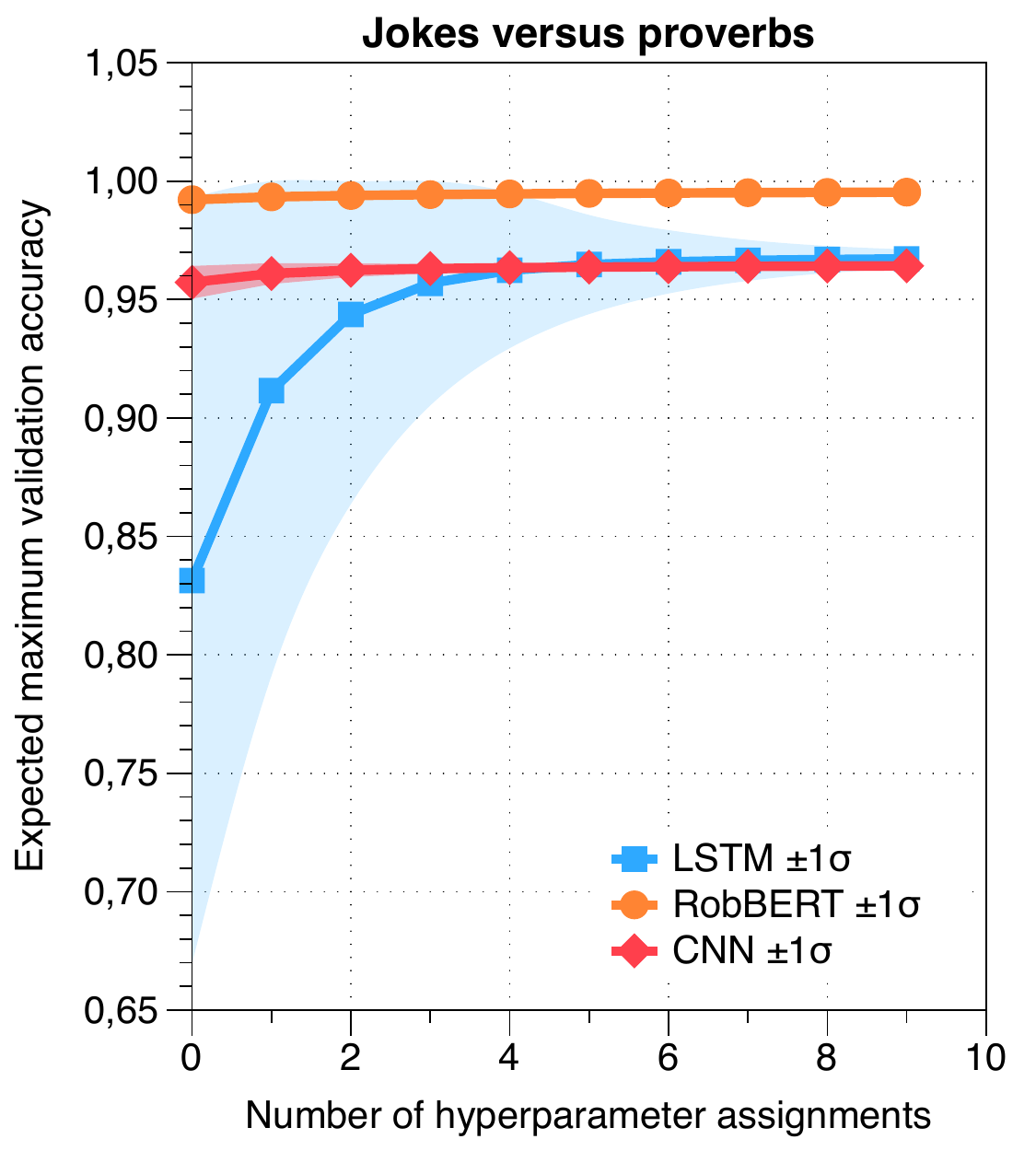} 
  \caption{Proverbs}
  \label{fig:sub-b}
\end{subfigure}
\begin{subfigure}{.245\textwidth}
  \centering
  \includegraphics[width=\linewidth]{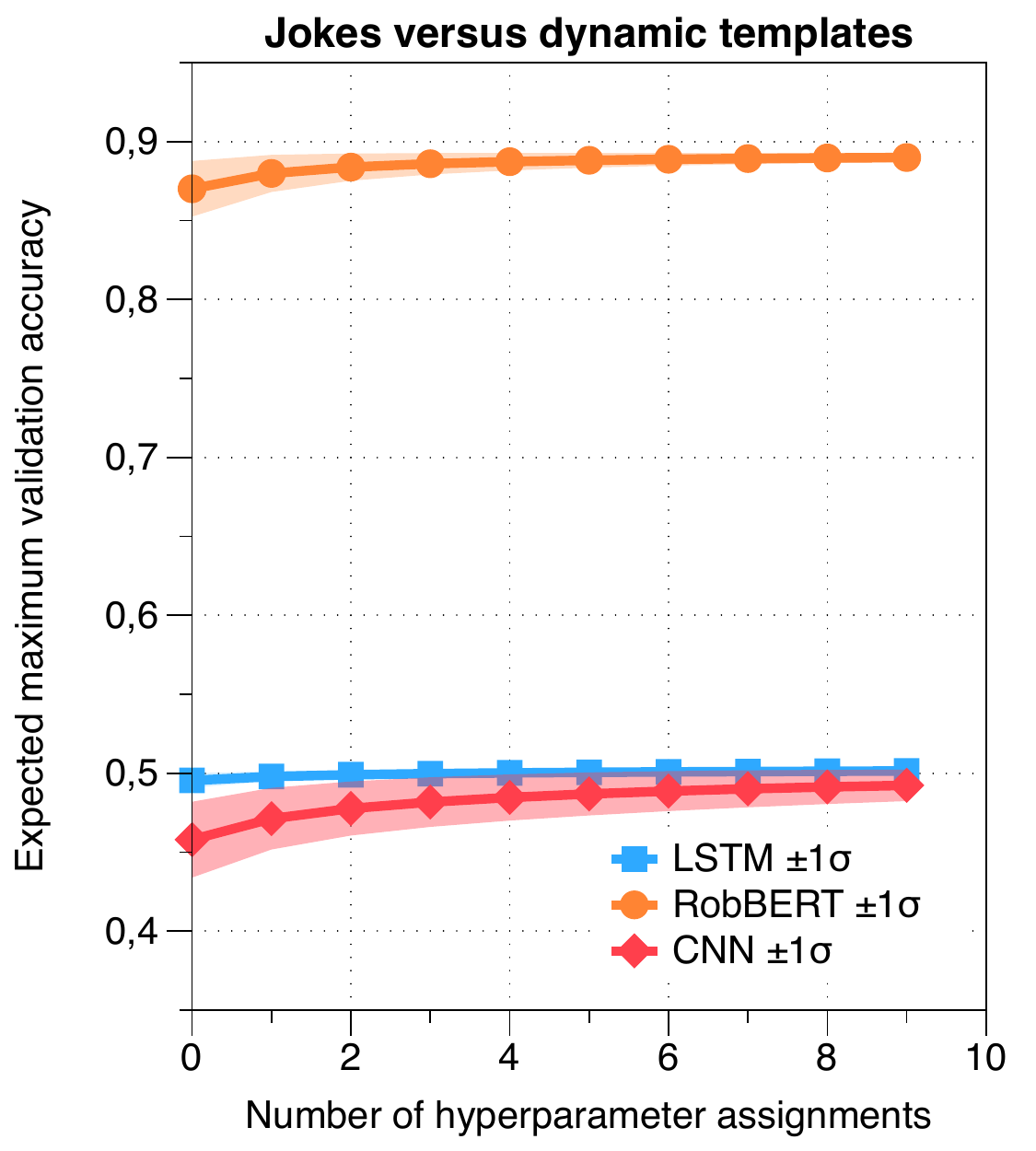} 
  \caption{Single DT}
  \label{fig:sub-c}
\end{subfigure}
\begin{subfigure}{.245\textwidth}
  \centering
  \includegraphics[width=\linewidth]{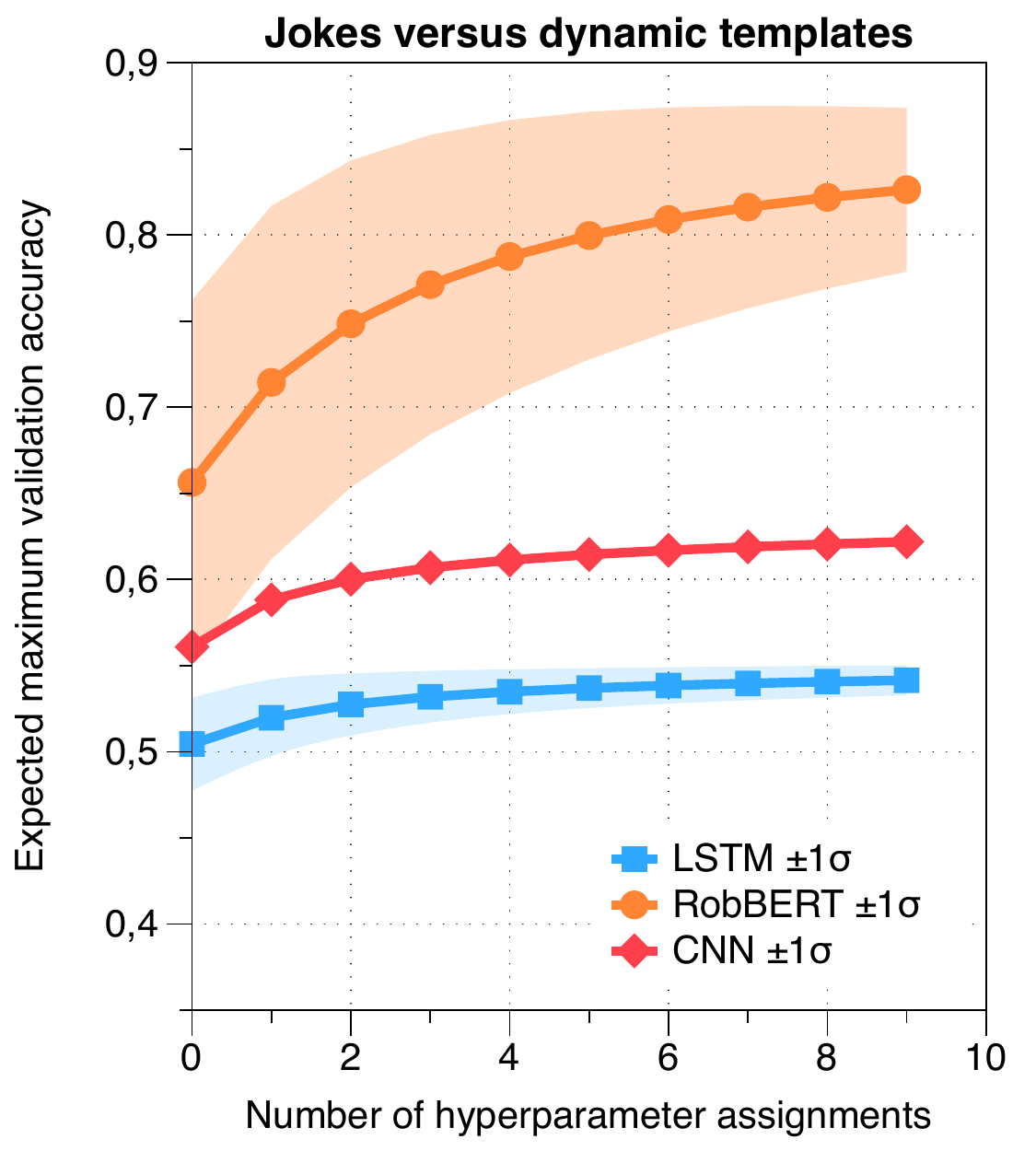} 
  \caption{Pairwise DT}
  \label{fig:sub-d}
\end{subfigure}
\caption{Estimated maximum validation accuracy~\cite{dodge2019show} in function of the number of hyperparameter trials for the LSTM, CNN, and RobBERT models.}
\label{fig:validation}
\end{figure}

In this binary classification experiment, the models classifies a given text as a joke or a non-joke.
We compared three different datasets, comparing jokes with news, with proverbs, and with generated jokes using the dynamic template algorithm.
We performed a random hyperparameter search with 10 runs for the LSTM, CNN, and RobBERT.
The full search space and other hyperparameters are listed in the Appendix in \autoref{tab:hp-space-lstm} and \autoref{tab:hp-space}.
In addition, we use these \emph{random} hyperparameter trials to estimate the maximum validation accuracy~\cite{dodge2019show}.
This allows us to compare performance without it being caused by a computational budget favoring one model.
\autoref{fig:validation} shows these estimates for the validation accuracy for all three datasets.
For both the news (\autoref{fig:sub-a}) and proverbs (\autoref{fig:sub-b}), both the CNN and the LSTM-based models perform a couple of percentage points below the RobBERT model.
More notably, the RobBERT model consistently achieves a validation accuracy around 99\%, whilst the LSTM has a higher variance than both the CNN model and RobBERT.
This indicates that the LSTM-based model is less robust to suboptimal hyperparameter assignment.

From these randomized trials, we select the best-performing model using the validation accuracy and evaluate on the held-out test set.
The results are presented in \autoref{tab:results-classification}.
The baseline, Naive Bayes, performs quite poorly, with the results being no better than random on all three datasets.
This is surprising, given that this method has been used successfully for humor detection in English for similar types of datasets, albeit often using handcrafted features instead of words~\cite{mihalcea2005humorrecognition,kiddon2011deviant,van2018homonym}.
This shows that a classifier using only token features is insufficient for all three Dutch humor datasets.
The LSTM and CNN models recognise about 94\% for the simple datasets
However, they both fail at distinguishing between jokes and non-jokes generated with dynamic templates.
This indicates that despite using Dutch word embeddings, these models likely still relies on vocabulary differences or the small lengths news and proverbs tends to have. %

Finetuning RobBERT gives us a testing accuracy of 98.8\% and 99.6\% on \emph{news} and \emph{proverbs}, respectively, and 89.2\% on the more challenging task with dynamic templates.
This shows that our newly created dataset is indeed more challenging than using non-jokes from completely different domains.
Interestingly, RobBERT's false positives contain many jokes with only limited replaced words, or that still made semantically coherent sense, e.g. the joke \emph{``Hoe heet de broer van Bill Mars? Bill Twix!''}\footnote{\emph{``What's the name of Bill Mars' brother? Bill Twix!''}}, only had one replacement (\emph{``Bruno'' to ``Bill'}), retaining some of the joke's humor.
RobBERT's higher performance than the other neural networks also illustrates the advantage of pre-trained language models for detecting semantic coherence in jokes, or at least distinguishing it from semantically incoherent non-jokes generated by the DT algorithm, either of which are useful properties.
To get more grasp on this, we classified all elements of the news and the proverbs dataset using the finetuned RobBERT model for the jokes versus dynamic template single setting. This input data was thus completely out-of-domain for this model.
We found that 93.23\% of the news and 73\% of the proverbs were labeled as a joke by this model, indicating that at least for such relatively short strings, the DT model might rely on topical or semantic coherence to recognise humor.

\begin{table}[tb]
\centering
\caption{Classification results for the held-out test set on three datasets versus the Jokes dataset. For the accuracy, we additionally report the 95\% CI.}
\label{tab:results-classification}

\begin{tabular}{@{}lcccccccc@{}}
\toprule
                                   & \multicolumn{2}{c}{\textbf{    }}                                     & \multicolumn{2}{c}{\textbf{        }}                                 & \multicolumn{4}{c}{\textbf{Dynamic template}}                           \\
                                    \cmidrule(l){6-9}                                                                                                                                                                      
                                   & \multicolumn{2}{c}{\textbf{News}}                                     & \multicolumn{2}{c}{\textbf{Proverbs}}                                 & \multicolumn{2}{c}{\textbf{Single}} & \multicolumn{2}{c}{\textbf{Pairwise}}                           \\
                                   \cmidrule(l){2-3} \cmidrule(l){4-5} \cmidrule(l){6-7} \cmidrule(l){8-9}
\multicolumn{1}{l}{\textbf{Model}} & \multicolumn{1}{c}{ACC {[}\%{]}} & \multicolumn{1}{c}{$F_1$ {[}\%{]}} & \multicolumn{1}{c}{ACC {[}\%{]}} & \multicolumn{1}{c}{$F_1$ {[}\%{]}} & \multicolumn{1}{c}{ACC {[}\%{]}} & \multicolumn{1}{c}{$F_1$ {[}\%{]}} & \multicolumn{1}{c}{ACC {[}\%{]}} & \multicolumn{1}{c}{$F_1$ {[}\%{]}}    \\ \midrule
Naive Bayes~~                      & $51.0\pm3.1$                     & $49.3$                             & $60.2\pm3.5$                     & $50.5$                             & $49.9\pm3.1$                     & $49.9$                     & - & -           \\
LSTM                               & $94.0\pm1.5$                     & $94.0$                             & $94.4\pm1.6$                     & $94.1$                             & $46.8\pm3.1$                     & $35.2$                     & $47.9 \pm 4.4$         & $32.4$               \\
CNN                               & $93.6\pm1.5$                     & $93.6$                             &94.1$\pm1.7$                     & $93.6$                             & $47.4\pm3.1$                     & $46.3$                     & $58.6\pm4.4$         & $58.5$               \\

RobBERT                            & $\mathbf{98.8}\pm0.7$                     & $\mathbf{98.8}$                             & $\mathbf{99.6}\pm0.4$                     & $\mathbf{99.6}$                             & $\mathbf{89.2}\pm1.9$                     & $\mathbf{89.1}$                     & $\mathbf{82.51} \pm 3.4 $       & $\mathbf{82.5}$             \\ \bottomrule
\end{tabular}

\end{table}

\subsection{Pairwise Classification Experiment}

We additionally perform an experiment where the joke and their non-joke counterpart, generated by the DT algorithm, are directly compared in a pairwise fashion.
The model thus has to recognise which one is more humorous than the other, opening the way for humor preference learning algorithms \cite{furnkranz2003pairwise}.

We evaluated an LSTM model with two separate recurrent layers with trainable Dutch word embeddings that are concatenated before a fully connected layer, which was also used for argument classification~\cite{delobelle2019computational}.
We evaluated a CNN model following a similar approach, with the same base architecture as in the previous experiment.
For RobBERT, we are using the same setup and hyperparameters, and feed the model both texts simultaneously, separated by the separator token.

In \autoref{tab:results-classification}, we can see that LSTMs are still unable to distinguish jokes from generated non-jokes, and CNNs only seeing a small performance boost in the pairwise case over the single case, illustrating the advantage of using such a challenging dataset.
RobBERT on the other hand is performing reasonably well but surprisingly loses some accuracy compared to the single classification case.
This is likely due to relatively more of the jokes being truncated to fit its input size limit, given that two texts are now fitted into the same input space.

\section{Future Work}

One way to improve the humor detection performance could be finding better ways of generating joke-like non-jokes, thus further increasing the difficulty of the dataset.
The DT algorithm is prone to occasional grammatical errors, which the models might pick up and use to just recognize grammatical errors, rather than recognize jokes.

These new humor detection algorithms also pave way for new humor generators, e.g. using a generate-and-test approach \cite{winters2019goofer}.
RobBERT could even fulfill two roles in such a generator, e.g. using a genetic algorithm that uses a pairwise joke detection head as tournament selection, and the word masking head to mutate the genomes.
Such a generator could also be useful in a collaborative setting where the humor comparator suggests better ways of phrasing a joke by subtly changing it e.g. by rearranging a potential punchline word to occur later.

\section{Conclusion}

We created three datasets for humor detection specifically for Dutch and proposed a new way to make more challenging humor detection datasets.
We hypothesized that currently popular approaches, like discerning news or proverbs, can rely on recognizing domain-specific vocabularies instead of the semantic coherence that makes jokes funny.
We illustrated this by constructing several models for humor detection on these new datasets; where we found that previous technologies indeed are not or barely able to distinguish jokes from similar non-jokes.
For a more modern architecture like RobBERT, the performance is only slightly lower for the generated non-jokes.
This shows that the generated negatives dataset is indeed more challenging, and that transformer models are a step in the right direction for humor detection given their context-sensitivity.
These datasets and findings open the way for interesting new, more context-aware Dutch joke detection and generation algorithms.

\subsection*{Acknowledgements}

Thomas Winters is a fellow of the Research Foundation-Flanders (FWO-Vlaanderen).
Pieter Delobelle was supported by the Research Foundation - Flanders under EOS No. 30992574 and received funding from the Flemish Government under the “Onderzoeksprogramma Artificiële Intelligentie (AI) Vlaanderen” programme.

\bibliographystyle{splncs04}
\bibliography{references}

\newpage
\appendix
\section{Dynamic Template parametrisation}
\label{app:dt}
The parameters used for the dynamic template algorithm to generate the non-jokes are a maximum word frequency of the 62\% percentile, and minimum number of replacement of at least one replacement for every 25 characters, and three randomly sampled jokes for context words

\section{Hyperparameter space}
\begin{table}[h]
\centering
\caption{The hyperparameter space used for the LSTM and CNN models with Dutch word embeddings.}
\label{tab:hp-space-lstm}
\begin{tabular}{@{}lll@{}}
\toprule        
\textbf{Hyperparameter}         & \textbf{LSTM model}                  & \textbf{CNN model}      \\ \midrule
adam\_epsilon                   & $10^{-8}$                       & $10^{-8}$           \\
fp16                            & False                           & False               \\
hidden\_dimension               & $i \in \{8, 16, 32, 64, 128\}$  & ---                 \\
learning\_rate                  & $[10^{-3}, 10^{-1}]$            & $[10^{-3}, 10^{-1}]$\\
pooling                         & ---                             & max                 \\
convolutional\_layers           & ---                             & 2                   \\
kernel\_size.                   & ---                             & 3                   \\
max\_grad\_norm                 & 1.0                             & 1.0                 \\
num\_train\_epochs              & 15                              & 15                  \\
batch\_size                     & 64                              & 64                  \\
seed                            & 1                               & 1                   \\
dropout                         & 0.1                             & 0.1                 \\ \bottomrule
\end{tabular}
\end{table}

\begin{table}[h]
\centering
\caption{The hyperparameter space used for finetuning RobBERT.}
\label{tab:hp-space}
\begin{tabular}{@{}ll@{}}
\toprule
\textbf{Hyperparameter}         & \textbf{Value}         \\ \midrule
adam\_epsilon                   & $10^{-8}$              \\
fp16                            & False                  \\
gradient\_accumulation\_steps   & $i \in \{1, 2, 3, 4\}$ \\
learning\_rate                  & $[10^{-6}, 10^{-4}]$   \\
max\_grad\_norm                 & 1.0                    \\
max\_steps                      & -1                     \\
num\_train\_epochs              & 3                      \\
per\_device\_eval\_batch\_size  & 8                      \\
per\_device\_train\_batch\_size & 8                      \\
seed                            & 1                      \\
warmup\_steps                   & 0                      \\
weight\_decay                   & $[0, 0.1]$             \\ \bottomrule
\end{tabular}
\end{table}

\end{document}